%% file: LA-MAML.tex
\documentclass[runningheads]{llncs}
\usepackage[T1]{fontenc}
\usepackage{cite}
\usepackage{graphicx}
\usepackage{textcomp}
\usepackage{xcolor}
\usepackage{algorithm}
\usepackage[noend]{algpseudocode}
\usepackage{amsmath,amssymb}
\floatname{algorithm}{Algorithm}
\newcommand{\Proc}[1]{\textbf{#1}}
\usepackage{booktabs} 
\usepackage{subcaption}
\usepackage{caption}
\usepackage{url}
\graphicspath{{figures/}}
\usepackage{multirow}
\usepackage{tabularx}
\usepackage{array}
\usepackage{makecell}

\usepackage{float}
\usepackage{balance}
\usepackage{setspace}
\usepackage{ragged2e}
\usepackage{placeins}
\def\BibTeX{{\rm B\kern-.05em{\sc i\kern-.025em b}\kern-.08em
    T\kern-.1667em\lower.7ex\hbox{E}\kern-.125emX}}
\begin{document}


\title{From Trajectories to Instructions: Language-Conditioned Meta-Reinforcement Learning}

\author{
Garvit Singla \thanks{Corresponding author: garvit.singla@iiitb.ac.in}
\and
Uma Maheswari Natarajan
\and \\
Raghuram Bharadwaj Diddigi \thanks{Dr. Raghuram Bharadwaj is supported by the Anusandhan National Research Foundation (ANRF) under the Prime Minister Early Career Research Grant ANRF/ECRG/2024/005235/ENS}
}

\institute{
International Institute of Information Technology Bangalore, India
}
\maketitle

\begin{abstract}
Model-Agnostic Meta-Learning (MAML) is a widely used framework for reinforcement learning (RL) that enables efficient transfer by learning global policy parameters that can be rapidly adapted to new tasks. MAML training proceeds in two loops: an inner loop where the global parameters are adapted to task-specific parameters, and an outer loop where these task-specific parameters are evaluated and losses are back-propagated to improve the global parameters. Traditionally, the inner loop adaptation is performed by collecting trajectories from the task environment and applying gradient updates on the empirical expected return, which can be a costly operation. We note that it is the outer loop that drives the actual learning of global parameters, and therefore the inner loop adaptation mechanism need not be restricted to be gradient-based. This observation leads us to ask: Can we replace the inner loop trajectory collection and gradient update with a simpler, task-specific signal? In many practical settings, tasks are naturally accompanied by language instructions. Leveraging these instructions as a direct task-specific signal, we propose LA-MAML (Language Adapted MAML), which modifies the inner loop by adapting the global policy parameters in a single step through a learned embedding of the task instruction, replacing the inner loop trajectory collection and gradient-based updates. Experiments on the BabyAI benchmark demonstrate that LA-MAML achieves competitive or improved performance compared to baselines at a significantly lower per-iteration wall-clock training time. These results demonstrate that language instructions are an effective and efficient substitute for trajectory-based inner loop adaptation in meta RL. 


\end{abstract}

\section{Introduction}
Reinforcement Learning (RL) \cite{RL} provides a principled framework for training agents to learn sequences of actions that maximize long-term reward in uncertain environments. When combined with neural networks, deep reinforcement learning (deep RL) \cite{mnih2013playing} has demonstrated success across a wide range of real-world domains, including autonomous driving \cite{bojarski2016end}, robotics \cite{levine2016end, ibarz2021train}, healthcare \cite{liu2022deep}, and strategic games \cite{gameofgo, humanlevelgame}. Despite these successes, deep RL methods are often computationally intensive and highly sensitive to hyperparameter choices, making training new tasks from scratch costly and brittle. Consequently, enabling agents to reuse knowledge acquired from prior tasks to accelerate learning on related tasks has become increasingly important. Transfer learning addresses this challenge by allowing RL agents to leverage previously learned representations and behaviors \cite{rusu2018meta}, thereby reducing data requirements and training time for new tasks. Such capability is particularly critical in practical settings where rapid adaptation is essential and repeated interaction with the environment is expensive or limited.

Model-Agnostic Meta-Learning (MAML) \cite{MAML} is a widely used transfer learning framework designed to enable rapid adaptation across tasks in both supervised learning and reinforcement learning settings. The central objective of MAML is to learn a global parameter initialization that can be easily adapted to related downstream tasks. In supervised learning, this adaptation is performed by minimizing the task-specific loss on the provided training examples, and the global parameters are subsequently optimized such that these adapted solutions perform well on held-out validation data. This process encourages the global parameters to lie in a region of the parameter space from which effective task adaptation is possible. During inference, the learned global parameters are adapted using a small number of examples from a previously unseen task to obtain a (possibly approximate) task-specific solution; even when not optimal, this adapted solution typically serves as a strong initialization.

In the reinforcement learning setting, task adaptation proceeds differently. The global policy parameters are adapted to the specific tasks using policy gradient updates computed from trajectories collected by interacting with the environment. These trajectories are generated by executing the policy corresponding to the current global parameters. The global parameters are then optimized such that the adapted policies achieve high return on their respective tasks. This interaction-based adaptation introduces an inherent trade-off during inference. Using a small number of trajectories leads to high-variance gradient estimates and unstable adaptation, while collecting more trajectories improves stability at the cost of increased computational and interaction overhead.

\begin{figure}[ht!]
\centering
      \includegraphics[scale = 0.5]{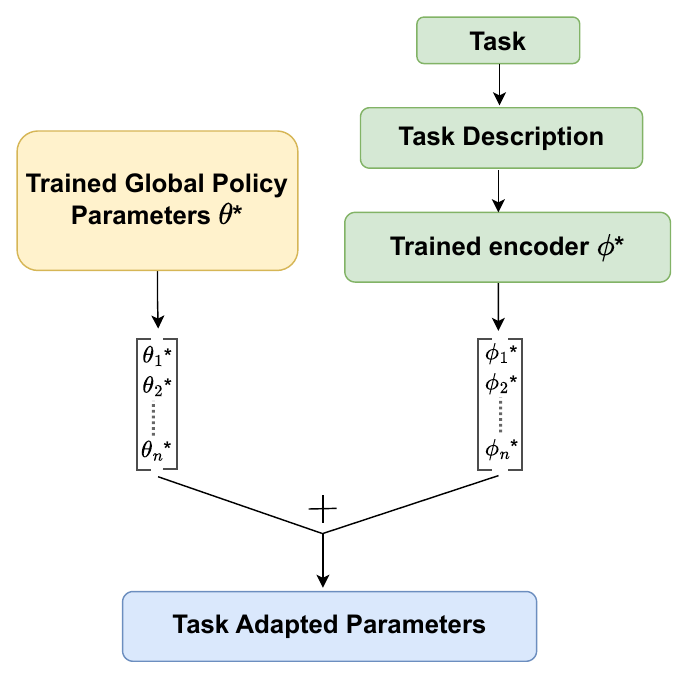}
\caption{Our Proposed LA-MAML trains the global parameters $\theta^*$ and trained encoder $\phi^*$ in an end-to-end fashion. During inference, the task adaption for task $\tau$ is obtained by simply adding $\theta^*$ and $\phi^*(\tau)$ }
\label{fig:proposed_approach_diagram}
\end{figure}

Motivated by the limitations of interaction-driven adaptation, we investigate whether a simpler and more reliable mechanism can be used to adapt global policy parameters to new tasks. In many practical reinforcement learning settings, tasks are naturally accompanied by high-level natural language descriptions that specify the task objectives. We leverage this structure and propose Language-Adapted MAML (LA-MAML), a meta-learning framework that uses task instructions as a consistent signal for adaptation. The core idea is to jointly train a shared global policy parameterization and a language encoder such that their combination yields a task-specific policy that is well aligned with the corresponding task. During inference, adaptation to a new task requires only computing the embedding of the task instruction (from the trained encoder) and combining it with the global parameters, resulting in an effective task-specific policy. An overview of this inference procedure is illustrated in Figure~\ref{fig:proposed_approach_diagram}. 

The key contributions of this work are summarized as follows:
\begin{enumerate}
\item We propose LA-MAML, a novel meta-learning framework for RL that leverages natural language task descriptions through a learned encoder–adapter mechanism for task-specific policy adaptation.
\item We empirically evaluate LA-MAML on the BabyAI benchmarks, showing faster convergence during training (wall-clock time per iteration) and improved performance during inference compared to other baselines.
\item We conduct ablation study to better understand the role of language in our proposed task adaptation.
\end{enumerate}

\section{Related Works}
In this section, we review prior literature by categorizing them into three key areas relevant to our work: efficiency in MAML, transfer learning extensions of MAML, and language-based task descriptions in meta-RL. 

\subsection{Efficiency in MAML}
In standard MAML, the framework incorporates two loops: an inner loop for adapting to individual tasks and an outer loop for generalization across a distribution of tasks. However, the outer-loop optimization requires differentiating through the inner-loop updates, introducing second-order gradients and increasing computational cost. Reptile \cite{nichol2018reptile} avoids this by using only first-order information by repeatedly sampling tasks, performing task-specific gradient updates, and moving the initialization toward the adapted weights of those tasks. 
Next, \cite{raghu2019rapid} investigates whether MAML’s effectiveness arises from rapid inner-loop adaptation or from feature reuse across tasks. The authors find that feature reuse is the dominant factor, leading to the development of ANIL (Almost No Inner Loop), which updates only the task-specific head while keeping the shared backbone fixed. 
In \cite{zintgraf2019cavia}, the CAVIA method partitions model parameters into global shared weights and low-dimensional context parameters, of which only the latter are updated per task. 
While methods such as ANIL, CAVIA, and WarpGrad \cite{flennerhag2019meta} explore ways to simplify or stabilize the inner loop by restricting adaptation to small parameter subsets, introducing context vectors, or re-parameterizing the gradient space, they continue to operate within the confines of gradient-based meta-optimization. In contrast, LA-MAML takes a different approach by removing the inner loop entirely and performing adaptation through language-conditioned parameter modulation, thereby reducing computational overhead.

\subsection{Transfer Learning Extensions of MAML}
While early variants of MAML primarily focused on improving computational efficiency within the inner loop, a complementary research direction has explored enhancing MAML’s transferability across tasks. 
MetaICL \cite{min2022metaicl} introduces a meta-training framework where large pretrained language models (LMs) are fine-tuned on a diverse collection of tasks to improve in-context learning. 
The work by \cite{deb2022boosting} explores whether meta-learning can enhance multi-task instructional learning (MTIL), where LMs are trained to follow natural language instructions across tasks. 
These studies extend MAML into the domain of large-scale language modeling, showing that meta-training across diverse tasks enhances in-context generalization. However, they still depend on gradient-based or data-driven meta optimization, where task adaptation is achieved implicitly through context or learned initialization. 

\subsection{Language-Based Task Descriptions in Meta-RL}
Finally, we highlight works showing how language can serve as a structured channel for task specification in meta-RL. 
The authors in \cite{bing2022meta} propose a meta-RL algorithm that uses natural language instructions with reward signals to guide adaptation across multiple manipulation tasks. 
The work by \cite{yao2022symmetry} introduces a dual-MDP formulation that incorporates both task symmetry and language instructions into meta-RL. 
Together with the imitation learning framework in \cite{stepputtis2020language}, these studies show that natural language provides an interpretable representation of task intent for guiding policy adaptation and improving generalization.
Building on this foundation, our proposed LA-MAML extends the integration of language by treating language embeddings as direct modulators of model parameters, enabling more efficient and interpretable task adaptation. 

\section{Background}
We first briefly describe the Model-Agnostic Meta-Learning (MAML) algorithm \cite{MAML}, followed by our proposed approach.

MAML is a widely used meta-learning framework designed to enable rapid adaptation to new tasks without retraining from scratch. It achieves this by learning an initialization of parameters $\theta$, that can be efficiently adapted to new tasks using only a few task-specific samples. 
To achieve this, MAML operates in two loops: inner-loop (task-specific adaptation) and outer loop (meta-update of global parameters). To formalize this process, tasks are sampled from a distribution $p(\mathcal{T})$ \cite{MAML}, which defines the task space considered during meta-training:
\begin{equation}
T_1, T_2, \dots, T_n \sim p(\mathcal{T}),
\label{eq:task_dist}
\end{equation}
where each task $T_i$ is associated with its own dataset. The model $f_\theta$, parameterized by $\theta$, corresponds to a policy $\pi_\theta(a|s)$ in reinforcement learning. For each task $T_i$, trajectories are collected using $\pi_\theta$
, which are used 
to compute the task-specific loss $L_{T_i}(\pi_\theta)$, defined as the negative expected return. The parameters are then adapted via gradient descent:
\begin{equation}
\theta'_i = \theta - \alpha \nabla_\theta L_{T_i}(\pi_\theta),
\label{eq:maml_inner}
\end{equation}
where $\alpha$ is the inner-loop learning rate.

After task-specific adaptation, the outer loop optimizes the initialization $\theta$ to generalize across tasks. The adapted parameters $\theta'_i$ define the policy $\pi_{\theta'_i}$, whose performance on task $T_i$ yields the loss $L_{T_i}(\pi_{\theta'_i})$. Aggregating these losses 
across tasks 
gives the meta-objective:
\begin{equation}
\mathcal{L}_{\text{meta}}(\theta) = \sum_{T_i \sim p(\mathcal{T})} L_{T_i}\!\left(\pi_{\theta'_i}\right),
\label{eq:meta_loss}
\end{equation}
which is used to update $\theta$:
\begin{equation}
\theta \leftarrow \theta - \beta \nabla_\theta \mathcal{L}_{\text{meta}}(\theta),
\label{eq:meta_update}
\end{equation}
where $\beta$ is the meta-learning rate. This enables the model to achieve strong performance on new tasks with minimal adaptation.

However, MAML relies on gradient-based adaptation requiring repeated trajectory collection, which becomes challenging in stochastic environments. Limited trajectories lead to high-variance gradient estimates and unstable policies, while increasing trajectories improves stability at the cost of higher interaction and computational overhead. This challenge persists both during meta-training and at inference, where adapting to new tasks requires environment interaction.

To address these limitations, we propose a framework that leverages natural language instructions as a structured representation of task objectives. By mapping language descriptions to parameter adjustments, our approach enables direct and efficient task adaptation without requiring trajectory collection or gradient-based inner-loop updates during adaptation. This reduces the computational burden associated with MAML while providing a more flexible mechanism for task adaptation.

\section{Proposed Methodology}
\label{PM}
We propose a novel meta-learning framework that accelerates task adaptation through a language-based setup, where task instructions are expressed in natural language and transformed into feature vectors. This approach bypasses the need for gradient-driven updates, which lie at the core of the conventional gradient-based procedure of MAML \cite{MAML}.

In our formulation, the adapted parameters $\theta'_i$, which define the policy $\pi_{\theta'_i}$ is obtained without explicit trajectory collection or gradient steps. Rather, they are derived through a direct offset, derived from natural language task descriptions:

\begin{equation}
\theta'_i = \theta + \delta f_\phi(\text{task}),
\label{eq:adapted_parameters}
\end{equation}
where $\delta$ is the step-size parameter,  $f_\phi(\text{task})$ is a learned function, parameterized by $\phi$, that maps the natural language description of a task into task-specific parameter offsets.

To realize this mapping function $f_\phi$, two integrated components are used. The first is a task encoder which transforms the natural language description into an embedding that captures the semantic intent of the task.
In our approach, this encoder is a pretrained Sentence Transformer \cite{reimers2019sentence} model (all-MiniLM-L6-v2).
The second is an adapter network that projects this embedding into the policy parameter space. It generates task-specific offsets that convert the high-level task representation into concrete parameter updates. These updates are then combined with the global initialization $\theta$ to yield the adapted parameters $\theta'_i$.

While our approach introduces novelty by replacing the gradient-based inner-loop with direct language driven parameter adaptation, the outer loop continues to function analogously to MAML. It optimizes the global initialization $\theta$ so that it remains broadly adaptable across tasks, while simultaneously training the language parameters $\phi$ of adapter network to generate effective task-specific offsets
from the frozen language embeddings. 
Through this joint optimization, the model not only preserves a generalizable initialization but also learns to exploit task descriptions for rapid adaptation.

Therefore, the meta-objective for the proposed framework is defined as:
\begin{equation}
\mathcal{L}_{\text{meta}}(\theta, \phi) 
= \sum_{T_i \sim p(\mathcal{T})} 
L_{T_i}\!\Big( \pi_{\theta + \delta f_{\phi}(T_i)} \Big),
\label{eq:meta_obj}
\end{equation}

where each loss $L_{T_i}$ is computed on  trajectories sampled from the adapted policy $\pi_{\theta + \delta f_{\phi}(T_i)}$, which results from merging the global initialization with the task-specific adjustment.

To realize this meta-objective, the outer loop applies a Trust Region Policy Optimization (TRPO)~\cite{schulman2015trust} based natural gradient update to the joint parameters $[\theta,\phi]$:

\begin{equation}
[\theta, \phi] \;\leftarrow\; [\theta, \phi] \;-\; \tilde{\nabla}_{\theta,\phi}\, \mathcal{L}_{\text{meta}}(\theta,\phi),
\label{eq:outer_update_phi}
\end{equation}
where $\tilde{\nabla}$ denotes the natural gradient computed under the TRPO constraint.

In this proposed framework, the model learns a broadly adaptable global policy initialization parameters $\theta$ and an effective task-encoding function $f_\phi$, enabling efficient adaptation to unseen tasks without explicit trajectory collection or gradient updates in inner-loop. During inference, the task-specific parameters of the new task $T_{\text{new}}$ is obtained as: $\theta'_{new} = \theta + \delta f_\phi(T_{\text{new}})$.
This allows the model to generalize quickly to novel tasks with minimal computational overhead. We present the complete pseudo-code of our proposed approach in Algorithm \ref{alg:Main Algo}.

\begin{algorithm}[!t]
\caption{Language Adapted Policy for MAML (LA-MAML)}
\small
\label{alg:Main Algo}
\begin{algorithmic}[1]
\Require Task distribution $p(\mathcal{T})$, hyperparameters (episodes $K$, weighting factor $\lambda$, discount factor $\gamma$, learning rate $\delta$)
\State Randomly initialize policy parameters $\theta$ and task-encoding parameters $\phi$
\State Initialize frozen pretrained Sentence Transformer encoder
\For{meta-iteration $=1$ \textbf{to} $N$}
    \State Sample set of tasks $T_i \sim p(\mathcal{T})$
  \For{each task $T_i$}
    \State $s_i$ $\gets$ natural-language instruction for $T_i$
    \State task embedding $e_i \gets$ encode $s_i$ into a dense
    \Statex \hspace{9em}semantic representation using the 
    \Statex \hspace{9em}Sentence Transformer encoder
    \State parameter offsets $\tilde f_i$ $\gets$ project $e_i$ into policy 
    \Statex \hspace
    {9em} parameter space $\theta$ using the adapter
    \Statex \hspace
    {9.5em}network
    \State task-specific parameters $f_\phi(T_i)$ $\gets$ reshape $\tilde f_i$ to  
    \Statex \hspace
    {9em}match the layer dimensions of $\theta's$ 
    \State Obtain task-adapted parameters $\theta'_i$ with no 
    \Statex \hspace{3em}inner-loop gradients
    \Statex \hspace{8em} $\theta'_i$ $\gets$ $\theta$ + $\delta f_\phi(T_i)$
    \State Collect $K$ trajectories using policy $\pi_{\theta'_i}$ and store them
    \Statex \hspace{3em}in buffer space $\mathcal{D}_i$ 
    \State Fit value function $V_\psi$ on trajectories in $\mathcal{D}_i$
    \State Compute temporal-difference residuals $\delta_t$ using $V_\psi$
    \Statex \hspace{8em} $\delta_t = r_t + \gamma V_\psi(s_{t+1}) - V_\psi(s_t)$
    \State Compute advantages $\hat{A}_t$ using GAE\cite{schulman2015high}:
    \Statex \hspace{8em} $\hat{A}_t = \delta_t + \gamma \lambda \hat{A}_{t+1}$
    \State \text {Compute task-specific loss } $L_{T_i}$
    \Statex \hspace{5em} $L_{T_i} \gets -\mathbb{E}_{(s,a)\sim\mathcal{D}_i}\big[\log \pi(a\mid s;\theta'_i)\,\hat{A}_t\big]$
  \EndFor
  \State \text {Compute Meta loss } 
  \Statex \hspace{2em} $\mathcal{L}_{\text{meta}}(\theta,\phi) \gets \sum_i L_{T_i}$
  \State \text Update global parameters using TRPO\cite{schulman2015trust} 
  \Statex \hspace{2em} $(\theta,\phi) \gets \Proc{TRPO}\big((\theta,\phi);\ \mathcal{L}_{\text{meta}}(\theta,\phi))$
\EndFor
\State \Return $(\theta,\phi)$
\end{algorithmic}
\end{algorithm}

Through the proposed framework, we demonstrate that a language trained model can leverage human-interpretable task descriptions, enabling flexible adaptation to tasks specified in natural language. Together, these advantages lead to a more efficient and versatile transfer approach.

\section{Experiments and Results}

We now begin by discussing the experiments and results obtained using our proposed LA-MAML framework, as outlined in the Proposed Methodology section \ref{PM}. The objective is to evaluate the effectiveness of language-driven adaptation in diverse task settings without relying on gradient-based inner-loop updates.

To this end, we adopt the BabyAI framework \cite{chevalier2018babyai}, a MiniGrid-based grid world where tasks are specified in natural language. The BabyAI environment comprises of connected rooms populated with multiple distractor objects where the agent must navigate and interact with objects to execute the given instruction correctly.
We evaluate our method on seven BabyAI environments: \texttt{GoToLocal}, \texttt{PickupDist}, \texttt{GoToObjDoor}, \texttt{GoToOpen}, \texttt{OpenDoor}, \texttt{OpenDoorLoc}, and
\texttt{OpenDoorsOrder}, covering a broad range of task compositions and difficulty levels.
BabyAI features sparse rewards setting where the agent receives a reward only upon successful task completion, with no intermediate feedback. 

The following subsections describe the experimental setup and results, and analyze the contribution of individual components of our proposed algorithm through an ablation study.

\subsection{Experimental Setup}
Each BabyAI environment exhibits distinct dynamics and can be instantiated in multiple configurations that vary in room size (S) and number of distractors (N).
For example, \texttt{GoToLocal} includes configurations such as \texttt{GoToLocalS5N2}, \texttt{GoToLocalS7N4}, and \texttt{GoToLocalS8N3} which progressively increase task complexity. For a given configuration, tasks are specified by language-based goals over target objects. 
Using this setup, for each environment we sample a set of training tasks while reserving a separate set of unseen tasks for evaluation. During evaluation, we test on the held-out tasks across multiple configurations.

For comparison, we consider three baselines: (i) standard MAML \cite{MAML}, which performs gradient-based meta-learning across tasks, (ii) a language-conditioned policy without meta-learning, which maps the state and task instruction \((s, \ell)\) directly to actions, similar to BabyAI framework \cite{chevalier2018babyai} and \emph{does not} perform inner or outer-loop updates, and (iii) ANIL (Almost No Inner Loop) \cite{raghu2019rapid}, a simplified variant of MAML where inner-loop adaptation updates only the task-specific head (final layer) while keeping the shared representation network fixed.
Further, all methods share the same environment observation space.
To ensure a fair comparison, we use identical network architectures and hyperparameters across all comparison methods. 

\subsection{Results and Discussion}
\begin{figure}[t]
    \centering
    \setlength{\tabcolsep}{1pt}
    \begin{tabular}{cccc}
        \includegraphics[width=0.245\textwidth]{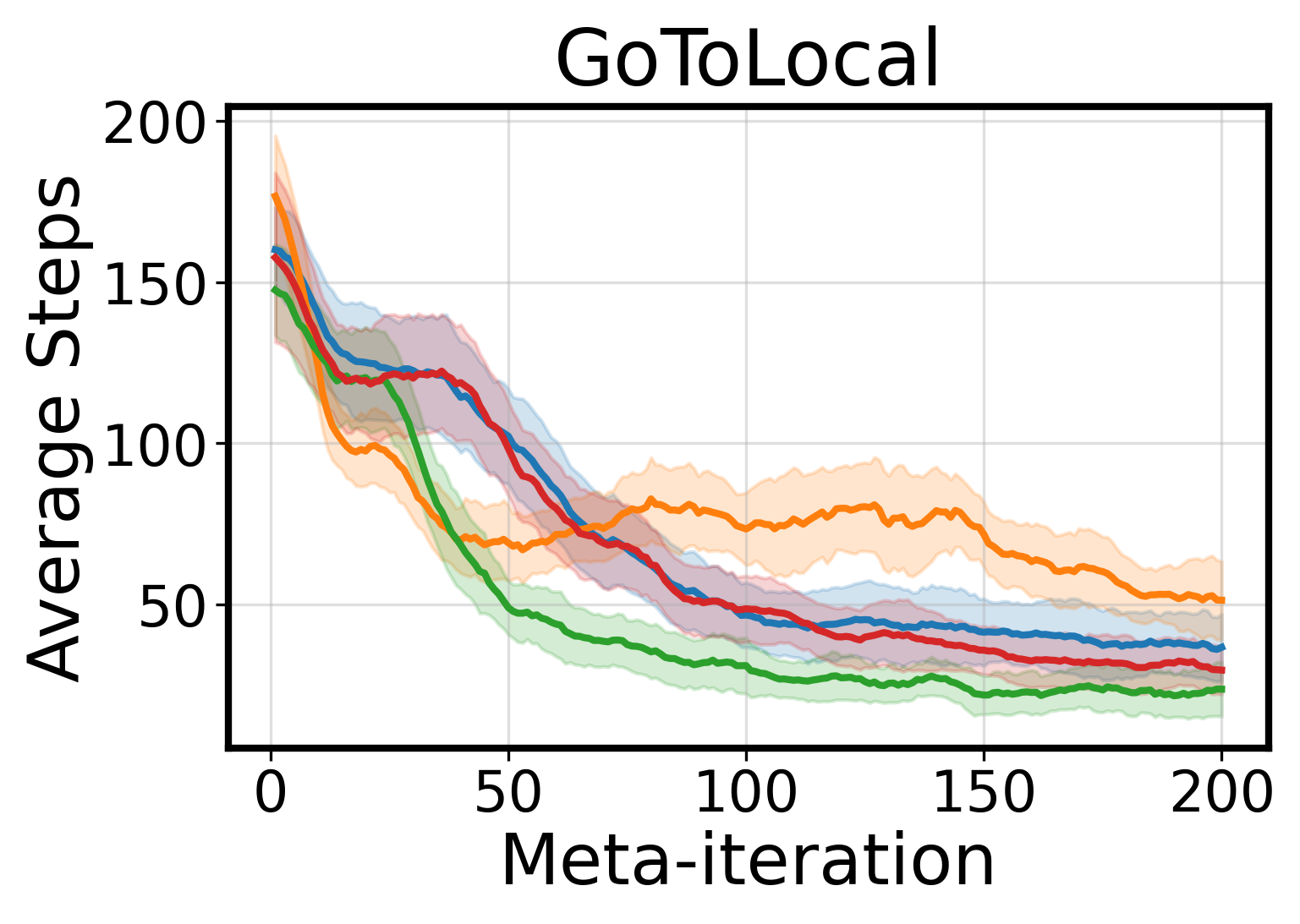} &
        \includegraphics[width=0.245\textwidth]{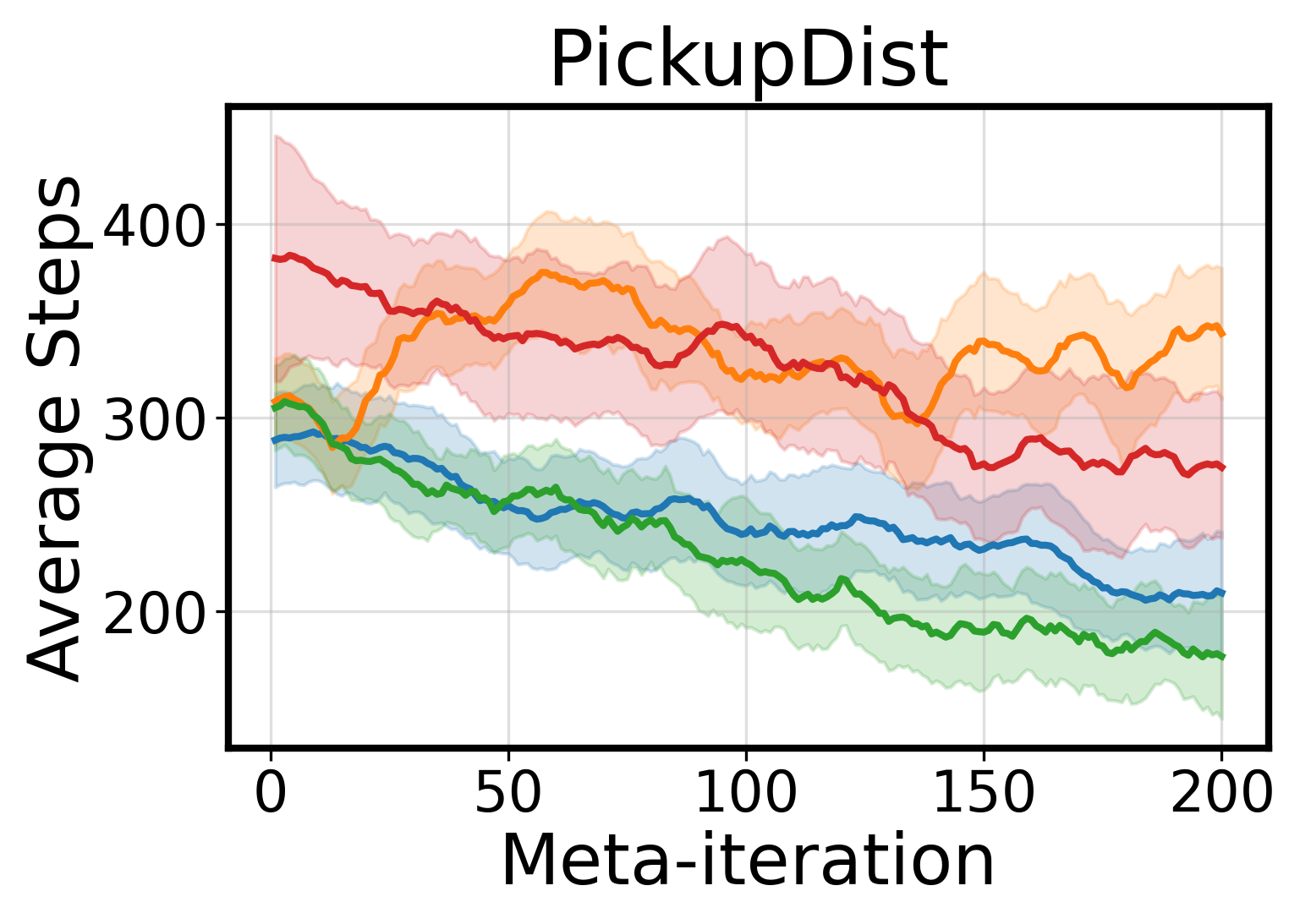} &
        \includegraphics[width=0.245\textwidth]{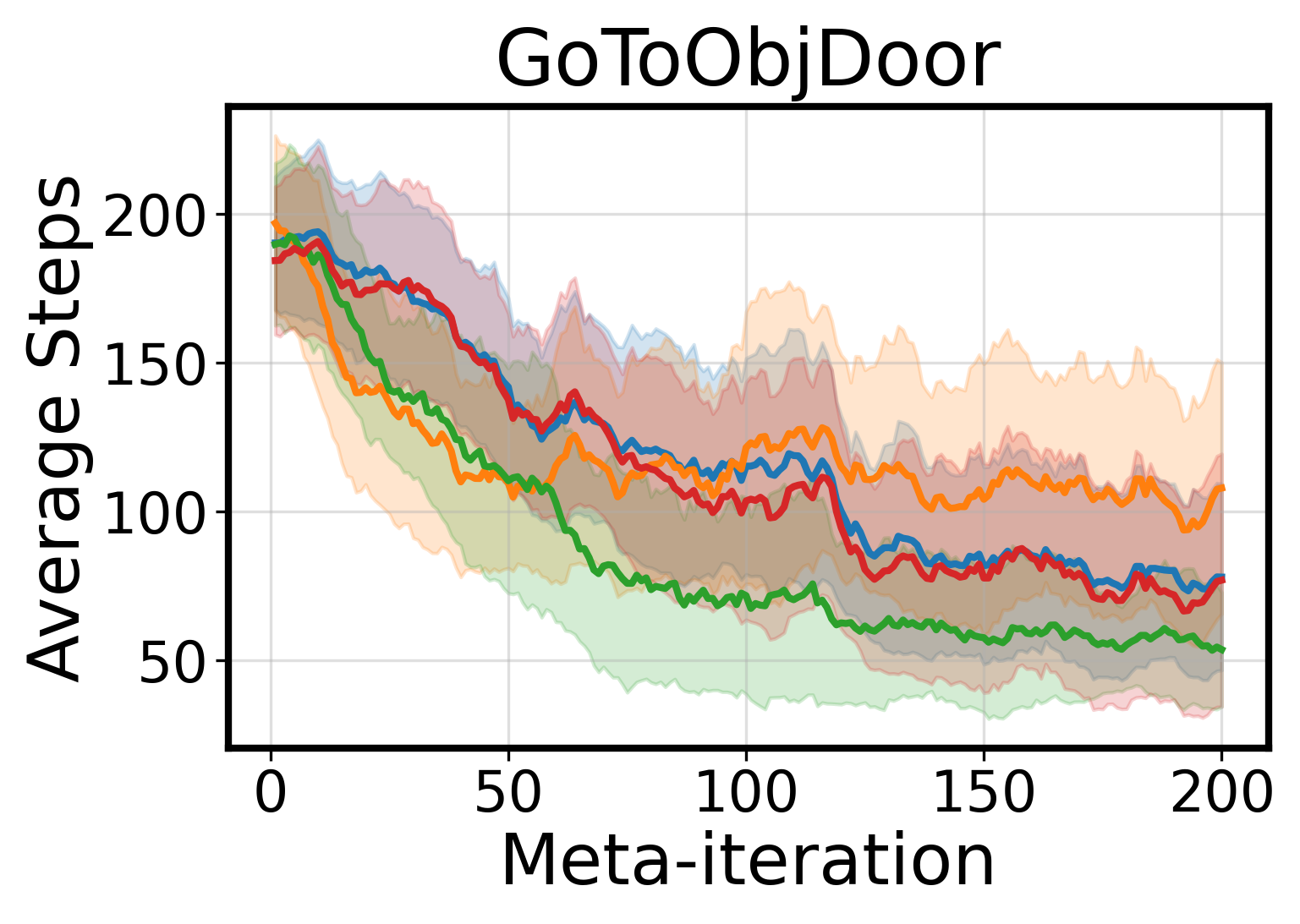} &        \includegraphics[width=0.245\textwidth]{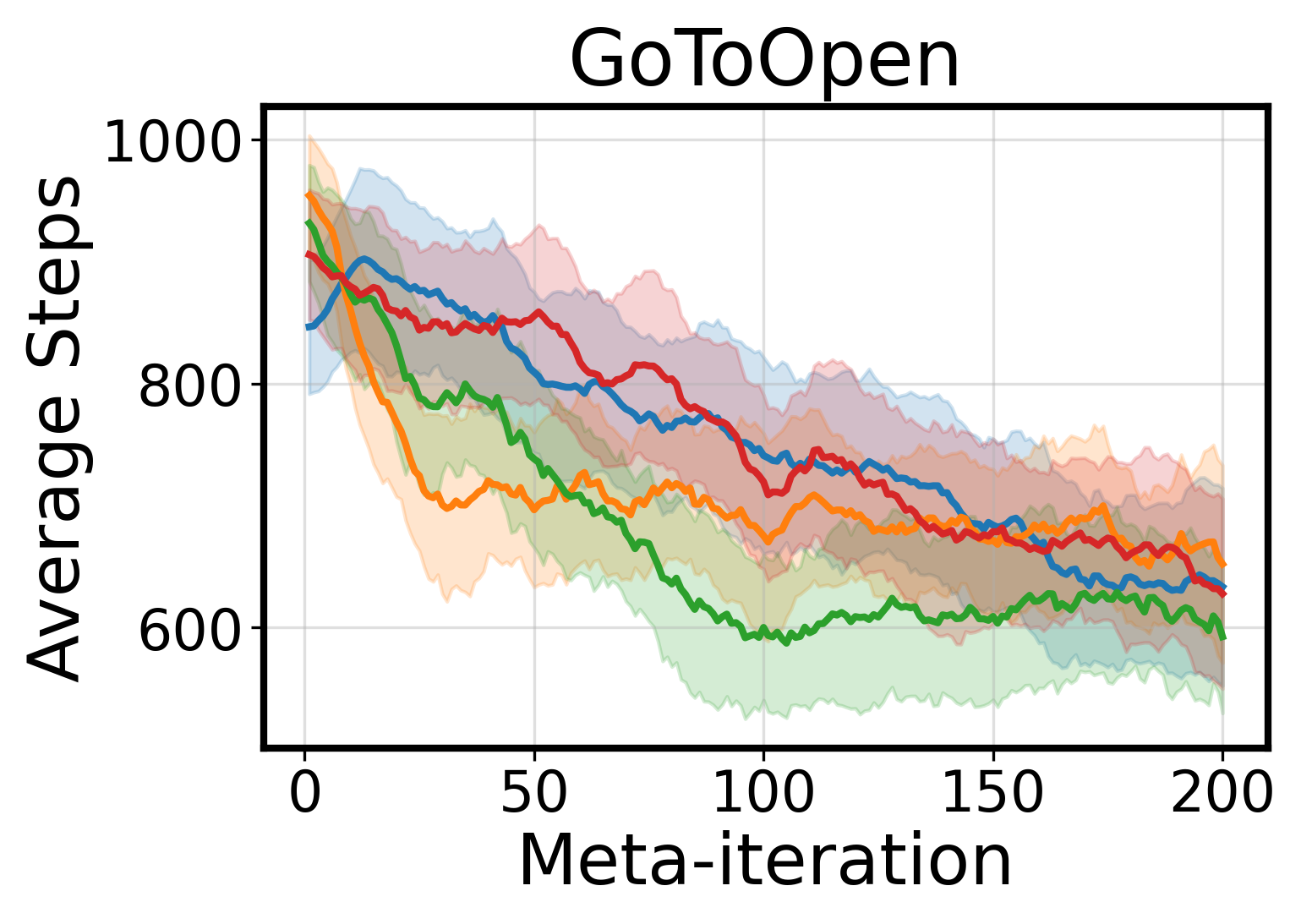}
         \\
  \end{tabular}
    \vspace{2pt}
    \begin{tabular}{ccc}
        \includegraphics[width=0.245\textwidth]{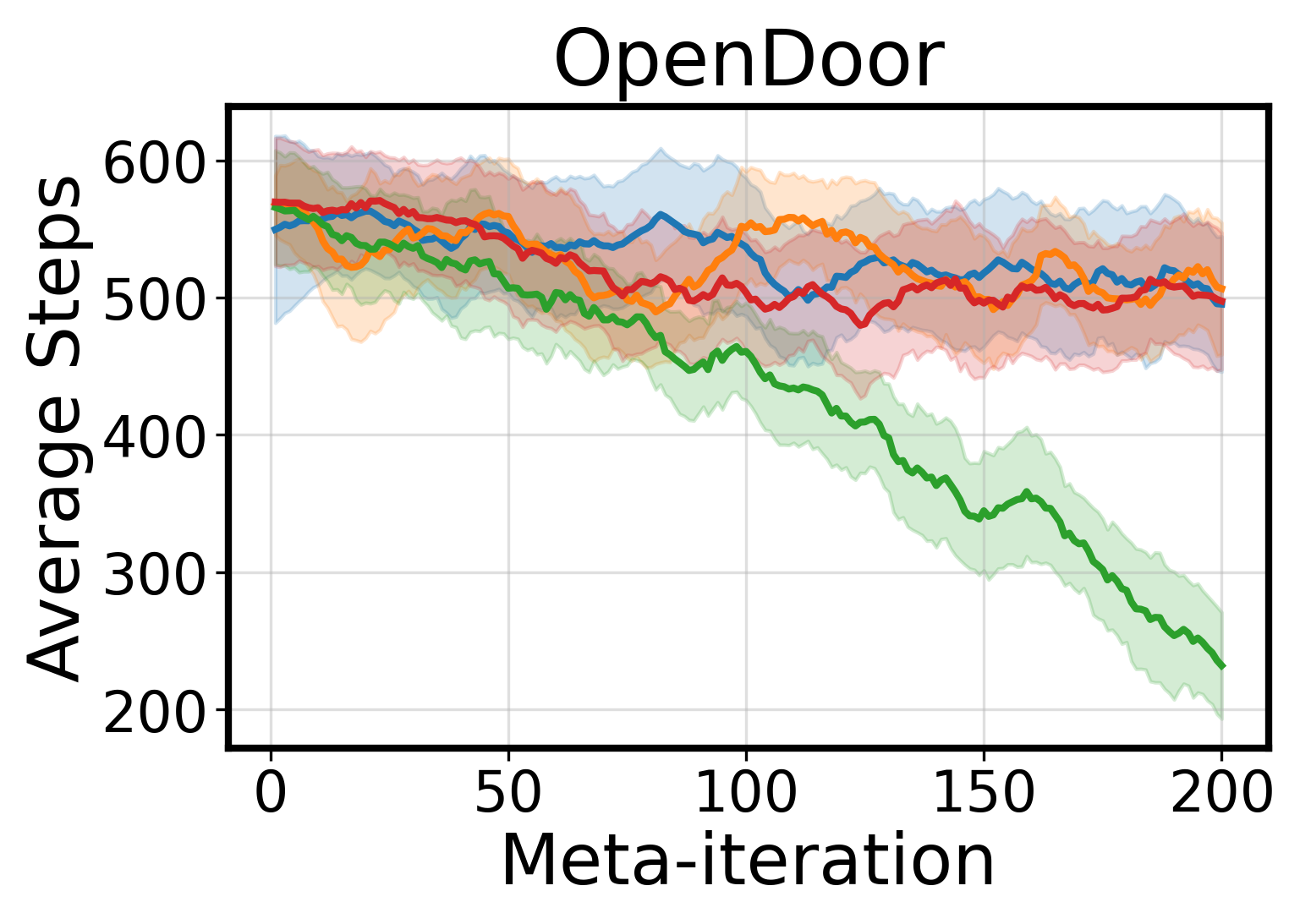} &
        \includegraphics[width=0.245\textwidth]{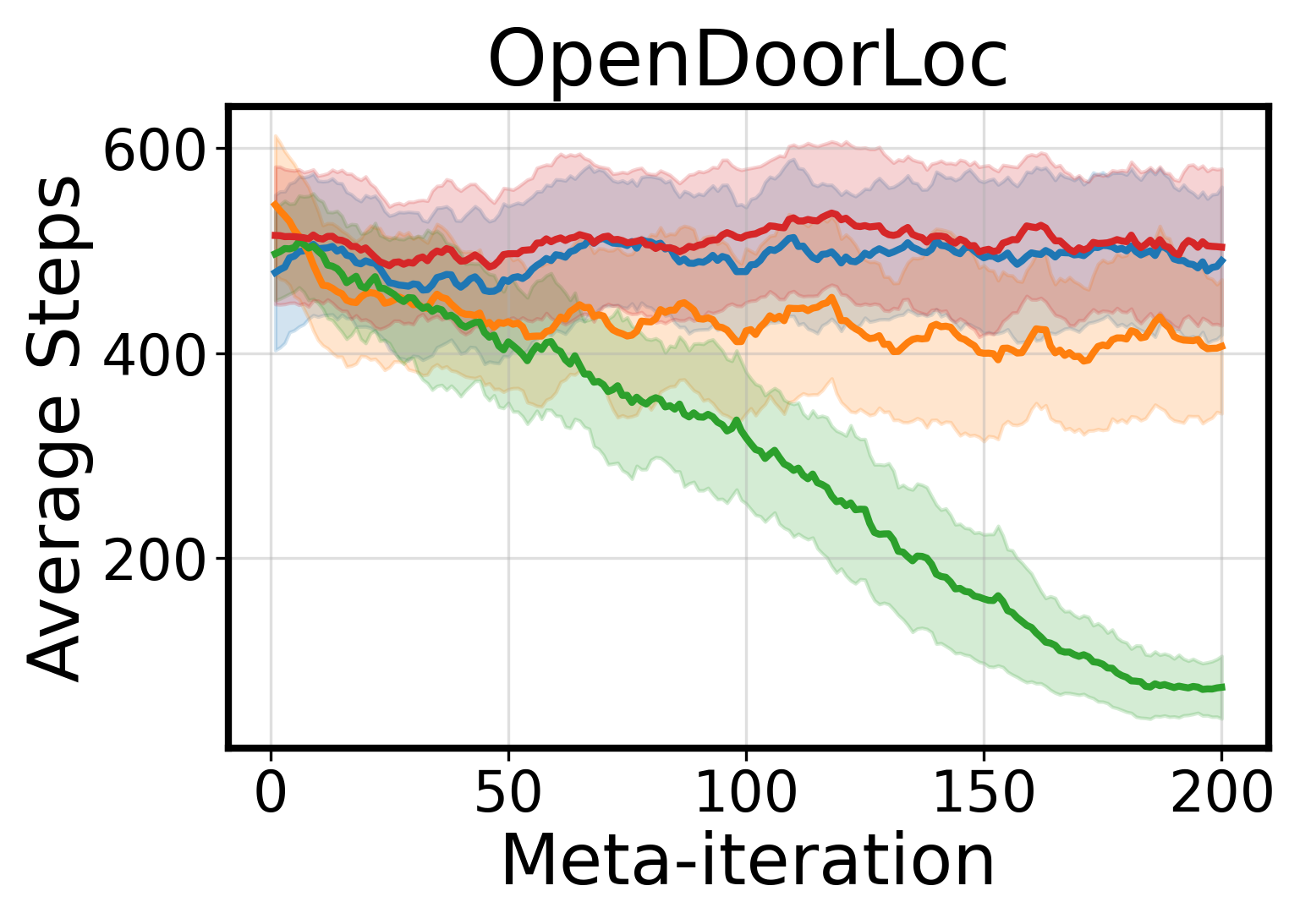} &
        \includegraphics[width=0.245\textwidth]{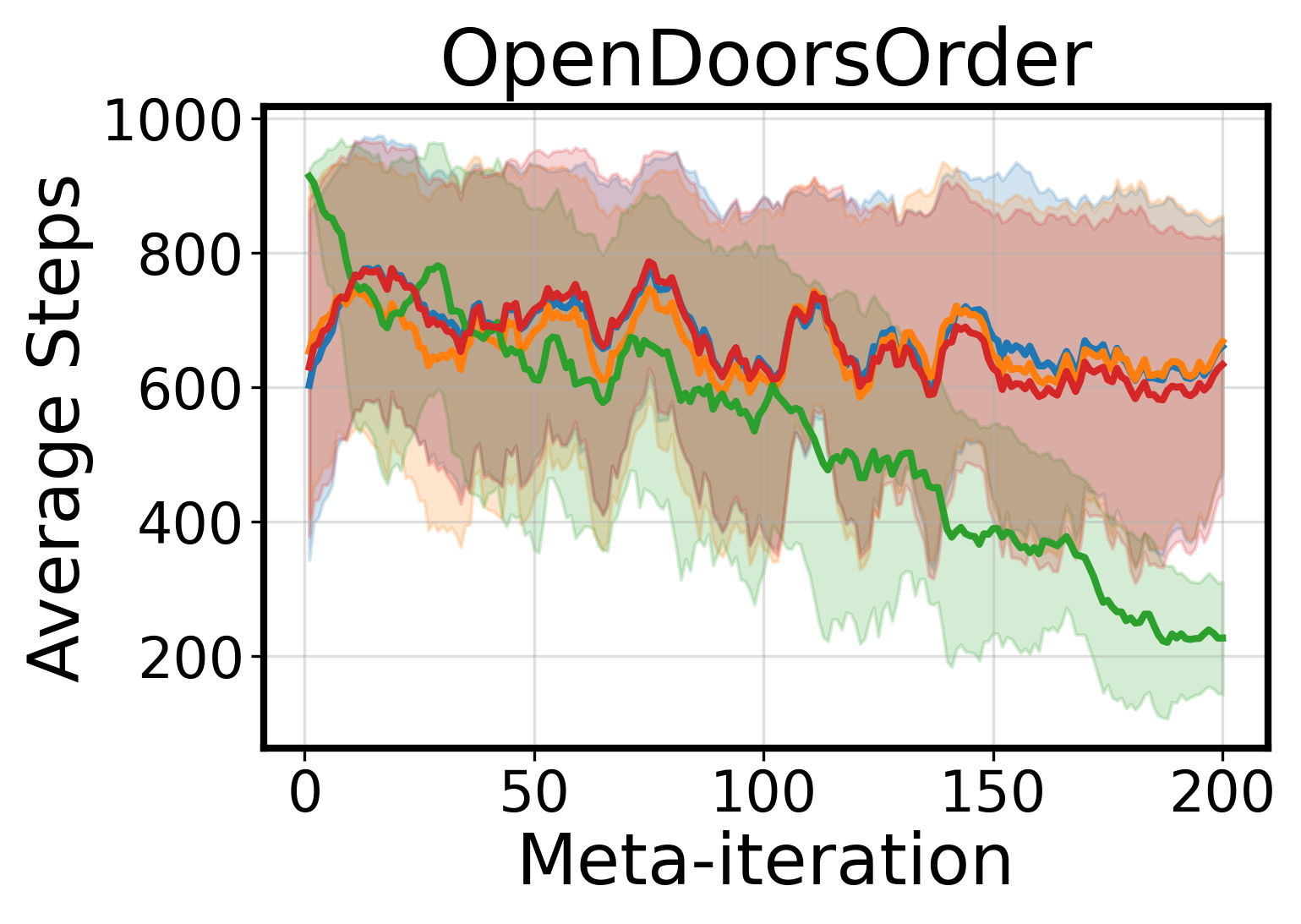}
        \\
    \end{tabular}
    \includegraphics[width=0.95\textwidth]{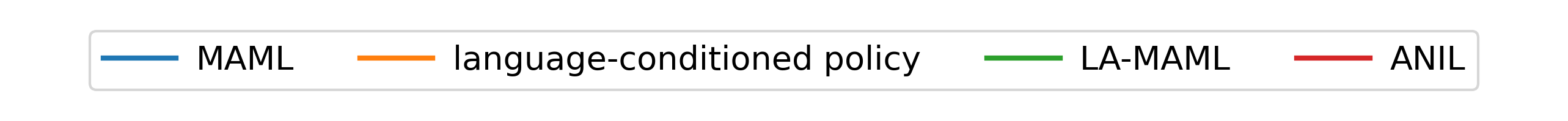}
    \caption{Training curves shows the convergence over meta-iterations for LA-MAML and baselines across seven environments. The shaded regions correspond to the standard deviation, while the solid lines represent the mean average steps per meta-iteration.}
    \label{fig:training_plots}
\end{figure}

\input{new_train_time}
\input{main_results}

Fig. 2 shows that LA-MAML consistently converges to lower average steps and at a faster rate compared to MAML, ANIL, and the language-conditioned policy baseline during the training. This demonstrates that replacing the gradient-based inner loop with direct language adaptation leads to more efficient training.

Further, from Tables 1, 2, we infer that:

\begin{enumerate}
    \item 
    LA-MAML achieves better evaluation performance while reducing wall-clock training time compared to gradient-based meta-learning baselines such as MAML and ANIL in most settings. Although it is not always the fastest method per iteration among all baselines, it provides a stronger balance between computational cost and adaptation performance.
    \item 
    The language-conditioned policy underperforms compared to LA-MAML on downstream tasks, indicating that language information alone is insufficient for effective adaptation. 
    Hence, improved performance of LA-MAML highlights the importance of meta-learning in enabling robust generalization across tasks.
\end{enumerate}

\vspace{-0.5em}
Overall, these results suggest that replacing the inner loop gradient update with a single step language-conditioned adaptation not only reduces per-iteration wall-clock training time but also yields competitive or improved task performance across environments
\footnote{The source code, hyperparameter settings and additional experimental results are available at: \url{https://github.com/garvitsingla/LA-MAML}} 

\subsection{Ablation Study}
We conduct an ablation study to further analyze the performance of our proposed approach. 

\input{ablation_during_training}
In this experiment, we analyze the impact of language during inference in the LA-MAML framework. The standard evaluation in LA-MAML uses both the learned global policy parameters $\theta$ and the language parameters $\phi$. Therefore during inference, we perform an ablation in which only the optimized global policy parameters are used, i.e., $\theta' = \theta$ so that the agent executes the task without relying on $\phi$. 

As shown in Table \ref{tab:ablation_untrained}, removing the language component results in a performance drop, indicating that the language parameters capture essential task-specific information during training that is later transferred and utilized at evaluation. This highlights that both $\theta$ and $\phi$ contribute significantly to the model's learning and execution: $\theta$ provides transferable knowledge, while $\phi$ imbues the policy with semantic context that guides task-specific adaptation.

\vspace{-0.5em}

\subsection{Limitations}
LA-MAML shows promising results in BabyAI settings, where templated instructions provide reliable task-relevant signals about the goal object, color, or navigation objective.
However, evaluating LA-MAML with more natural and potentially imperfect language, where task descriptions may be less structured, incomplete, or noisy, remains an important future direction.

\section{Conclusion}
In this work, we propose LA-MAML, a novel meta-reinforcement learning framework that leverages natural language as task descriptors to enable efficient and interpretable task adaptation. Unlike standard MAML, which depends on gradient-based inner-loop updates and trajectory collection for each new task, LA-MAML performs direct parameter adaptation driven by language semantics. 
This allows the model to infer task-specific behavior directly from linguistic cues, leading to faster and more efficient adaptation. 
The experiments across seven benchmark BabyAI environments demonstrate that language-driven adaptation achieves strong performance and computationally more efficient compared to both gradient-based meta-learning and a language-conditioned policy.

\bibliographystyle{splncs04} 
\bibliography{references} 

\end{document}

%% file: new_train_time.tex
\begin{table}[ht]
\centering
\small
\caption{Time taken per iteration during training (in seconds).}
\label{tab:env_training_time}
\setlength{\tabcolsep}{6pt}
\renewcommand{\arraystretch}{1.1}
\begin{tabular}{lcccc}
\hline
\textbf{Environment} &
\textbf{LA-MAML} &
\textbf{MAML} &
\makecell[c]{\textbf{Language}\\\textbf{cond. Policy} \\} &
\textbf{ANIL} \\
\hline
GoToLocal     & 5.31  & 7.42      & 2.53  & 7.00  \\
PickupDist    & 8.04  & 12.47     & 19.01 & 21.21 \\
GoToObjDoor   & 6.63  & 13.17     & 7.51  & 12.93 \\
GoToOpen      & 26.04 & 40.73     & 22.19 & 26.85 \\
OpenDoor      & 10.02 & 25.02     & 13.27 & 17.01 \\
OpenDoorLoc   & 13.39 & 24.94     & 24.43 & 19.82 \\
OpenDoorsOrder & 19.91  & 39.73     & 35.86  & 33.23 \\
\bottomrule
\end{tabular}
\end{table}

%% file: main_results.tex
\begin{table*}[htbp]
\centering
\small 
\caption{Performance comparison of LA-MAML with standard MAML, Language-conditioned Policy and ANIL across environments on new tasks using average steps (Mean $\pm$ Standard Deviation), including both successful and failed episodes, where maximum steps are capped as per environment dynamics (Lower is better).}
\label{tab:main_results}
\setlength{\tabcolsep}{3pt}
\renewcommand{\arraystretch}{1.15}
\begin{tabular*}{\textwidth}{@{\extracolsep{\fill}} l c c c c}
\toprule
\textbf{Environment} &
\textbf{LA-MAML} &
\textbf{MAML} &
\makecell[c]{\textbf{Language}\\\textbf{conditioned} \\\textbf{Policy}\\} &
\textbf{ANIL} \\
\midrule
GoToLocal
& \mbox{$\mathbf{43.59 \pm 22.35}$}
& \mbox{$44.19 \pm 17.27$}
& \mbox{$73.18 \pm 28.05$}
& \mbox{$47.43 \pm 20.41$} \\
PickupDist
& \mbox{$\mathbf{194.15 \pm 62.03}$}
& \mbox{$286.69 \pm 78.11$}
& \mbox{$358.39 \pm 68.66$}
& \mbox{$256.12 \pm 59.97$} \\
GoToObjDoor
& \mbox{$\mathbf{54.86 \pm 22.65}$}
& \mbox{$64.40 \pm 28.22$}
& \mbox{$87.55 \pm 39.54$}
& \mbox{$65.35 \pm 30.39$} \\
GoToOpen
& \mbox{$\mathbf{579.48 \pm 149.74}$}
& \mbox{$638.07 \pm 137.86$}
& \mbox{$659.26 \pm 147.80$}
& \mbox{$635.25 \pm 142.82$} \\
OpenDoor
& \mbox{$\mathbf{204.28 \pm 74.65}$}
& \mbox{$560.73 \pm 100.61$}
& \mbox{$463.42 \pm 90.21$}
& \mbox{$561.06 \pm 95.53$} \\
OpenDoorLoc
& \mbox{$\mathbf{135.58 \pm 91.36}$}
& \mbox{$516.08 \pm 115.41$}
& \mbox{$418.86 \pm 110.11$}
& \mbox{$487.98 \pm 117.36$} \\
OpenDoorsOrder
& \mbox{$\mathbf{317.59 \pm 172.37}$}
& \mbox{$553.73 \pm 272.87$}
& \mbox{$568.50 \pm 267.76$}
& \mbox{$514.78 \pm 251.16$} \\
\bottomrule
\end{tabular*}
\end{table*}

%% file: ablation_during_training.tex
\begin{table}[ht]
\centering
\small
\caption{Comparison of LA-MAML against inference with global parameters $\theta$ only. (Lower is better)}
\label{tab:ablation_untrained}
\setlength{\tabcolsep}{4pt}
\renewcommand{\arraystretch}{1.1}
\begin{tabular}{lcc}
\hline
\textbf{Environment} & \textbf{LA-MAML} &
\makecell[c]{\textbf{Inference using} \\ \textbf{$\theta$ only}} \\
\midrule
GoToLocal       & $44.32 \pm 26.49$   & $59.27 \pm 24.05$    \\
PickupDist      & $202.48 \pm 54.99$  & $263.88 \pm 49.16$   \\
GoToObjDoor     & $59.84 \pm 28.86$   & $83.38 \pm 40.82$    \\
GoToOpen        & $576.11 \pm 132.14$ & $658.25 \pm 174.74$  \\
OpenDoor        & $193.90 \pm 60.35$  & $450.68 \pm 101.08$  \\
OpenDoorLoc     & $141.93 \pm 96.33$  & $435.60 \pm 133.08$  \\
OpenDoorsOrder  & $299.52 \pm 179.98$ & $620.69 \pm 258.27$  \\
\hline
\end{tabular}
\end{table}